\pdfoutput=1

\documentclass[11pt]{article}

\usepackage[preprint]{coling}

\usepackage{times}
\usepackage{latexsym}

\usepackage[T1]{fontenc}

\usepackage[utf8]{inputenc}

\usepackage{microtype}

\usepackage{inconsolata}

\usepackage{dialogue}
\usepackage{enumitem}
\usepackage{makecell}
\usepackage{graphicx}
\title{Investigating the effect of Mental Models in User Interaction with an Adaptive Dialog Agent}

\author{Lindsey Vanderlyn
    \quad Dirk Väth
    \quad Ngoc Thang Vu
    \\
  University of Stuttgart \\
  Institute of Natural Language Processing \\
  \texttt{vanderly@ims.uni-stuttgart.de} \\
}

\begin{document}
\maketitle
\begin{abstract}
Mental models play an important role in whether user interaction with intelligent systems, such as dialog systems is successful or not.
Adaptive dialog systems present the opportunity to align a dialog agent's behavior with heterogeneous user expectations.
However, there has been little research into what mental models users form when interacting with a task-oriented dialog system, how these models affect users' interactions, or what role system adaptation can play in this process, making it challenging to avoid damage to human-AI partnership.
In this work, we collect a new publicly available dataset for exploring user mental models about information seeking dialog systems.
We demonstrate that users have a variety of conflicting mental models about such systems, the validity of which directly impacts the success of their interactions and perceived usability of system.
Furthermore, we show that adapting a dialog agent's behavior to better align with users' mental models, even when done implicitly, can improve perceived usability, dialog efficiency, and success.
To this end, we argue that implicit adaptation can be a valid strategy for task-oriented dialog systems, so long as developers first have a solid understanding of users' mental models.
\end{abstract}
\section{Introduction}

Adapting dialog agents' behavior to users has long been an area of interest in dialog research.
To this end, researchers have explored various strategies, e.g., language style \cite{MA202050}, sense of humor \citep{ritschel2018shaping}, recommendations \citep{he2023conversation}, etc.
The basic assumption underlying this research is that such changes to a dialog agent's behavior will lead to more successful or enjoyable interactions with users. 
However, this is not guaranteed.
\citet{bansal2019updates} and \citet{10.1145/3290605.3300714} experimented with human-AI and human-dialog system interaction respectively, and found that adaptation can actually damage human-agent partnership when done in a way which does not align with users' expectations.
Therefore, in order to develop successful adaptive dialog agents, it is important to first have a good understanding of what these user expectations are.

One way of understanding these is to measure user mental models. 
Mental models refer to a person's cognitive representation for how or why they believe a complex system (e.g., a dialog system) works \cite{johnson1980mental,halasz1983mental,norman2014some}. Rather than trying to process all details of such systems at once, users will create a simplified representation of the system in their mind~\cite{clements2004perspective}. 
These models can be arbitrarily simple -- e.g., \textit{``the dialog system can recognize keywords''} -- or complex -- e.g., \textit{``I would expect the chatbot to be able to answer simple questions, where they can retrieve the answers from my account information''} -- depending on the user.
These expectations and abstractions are shaped both through interaction with the system and through previous experience \citep{10.1145/3170427.3180286, Rutjes2019AIHCI}.
Research in human-computer interaction and human-centered AI has found that users' mental models of a system play an important role in predicting how they will interact with it \citep{10.1145/3290605.3300714} and that accurate mental models generally lead to more successful interactions \citep{10.1145/2207676.2207678, bansal2019beyond}.

Previous research into mental models of task-oriented dialog systems generally focuses either on single turn agents, like personal assistants \citep{10.1145/2858036.2858288, 10.1145/3170427.3180286,9538817}, or on collaborative game settings \citep{10.1145/3313831.3376316, weitz-etal-2021-fault}.
However, these domains only represent a small subset of dialog interactions.
Additionally, they do not consider the potential of adaptation to influence such interactions.
While there have been a limited number of studies looking at mental models in other domains \citep{10.1145/3411764.3445645,brachman2023follow}, to our knowledge, the only research exploring users' mental models around adaptive dialog agents was performed by \citet{10.1145/3290605.3300714}.
In their work, the researchers focused specifically on the scenario of users actively trying to teach an adaptive agent, with the assumption that an implicitly adaptive agent could be poorly accepted by users.
However, as such a teacher role imposes an additional cognitive load on users, the goal of this paper is to explore how users perceive and react to implicitly adaptive, task-oriented dialog agents.
Concretely, we investigate the following research questions:

\vspace{2pt}
\noindent\textbf{RQ1:} What role do users' mental models play in task-oriented dialog?
    \begin{itemize}[noitemsep,topsep=0pt]
        \item \textbf{RQ1.1:} What mental models do users have about task-oriented dialog systems before starting an interaction?
        \item \textbf{RQ1.2:} How do these mental models affect interaction with a dialog system?
    \end{itemize}
\noindent\textbf{RQ2:} What role does adaptation have on user mental models and interaction?
    \begin{itemize}[noitemsep,topsep=0pt]
        \item \textbf{RQ2.1:} How does interaction with an adaptive dialog system change user mental models compared to interaction with non-adaptive baselines?
        \item \textbf{RQ2.2:} Does implicit adaptation in line with user mental models affect success, trust, and usability?

\end{itemize}

To answer these questions, we implement three different types of task-oriented dialog system in the domain of business travel.
We then recruit 66 participants to take part in a user evaluation.
Each user interacts with either an implicitly adaptive dialog agent based on \citep{vath-etal-2023-conversational}, or one of two non-adaptive baselines, representing both extremes of the adaptive agent's behavioral spectrum.
We probe users for their mental models before and after they interact with the dialog system, analyzing how these affected their interaction, how their mental models were updated through the interaction, and how successful each interaction was.

Our main contributions are: 
1) Demonstrating that users have a wide variety of (contradictory) expectations for how to interact with an information-seeking dialog system. 
2) Showing users' mental models have a significant effect on how they interact with a dialog system, highlighting the need for adaptive systems.
3) Demonstrating that implicit adaptation can be done without negatively impacting mental models, and may in some cases improve user expectations.
4) Showing that implicit adaptation in line with users' mental models is not only rated more usable, compared to non-adaptive baselines, but also significantly increases dialog success.
5) Creating a new, publicly available dataset for studying mental models consisting of collected dialogs augmented with self-reported mental models (pre- \& post interaction) and logs of each user's interactions.

\section{Related Work}

\subsection{Adaptive Dialog Systems}
Research into adaptive dialog systems aims to align their behavior to users in order to improve interaction experience.
This research can focus on text-level adaptions or even adapting the agent's underlying behavior.
Text-level adaptation includes changing the dialog agent's linguistic style, e.g., adjusting the chatbot's utterances to match a user's emotional state \citep{MA202050}, personality \citep{doi:10.1137/1.9781611975321.71, 9801557}, or even adapting the complexity of language to a user's domain familiarity \citep{janarthanam-lemon-2014-adaptive}.
Behavioral adaptation approaches may rely on additional social cues, e.g., laughter \citep{ritschel2018shaping}, requiring users to intentionally fine-tune the dialog system's behavior \cite{chen2012critiquing, narducci2018improving}, or try to implicitly intuit cues for adaptation directly from the users' behavior \cite{vath-etal-2023-conversational}.

Trying to implicitly adapt an intelligent agent's behavior without a good understanding of users' mental models, however, can lead to a mismatch in user expectations and lead to less successful human-agent interactions \citep{weld2003automatically, bansal2019updates} and confusion \cite{10.1145/1240624.1240843}.
At the same time, asking users to directly provide feedback during the interaction interrupts the experience, which can make the dialog less natural.
Furthermore, asking for such feedback places additional cognitive load on the user, which is contrary to the goal of intelligent systems, namely to reduce a user's cognitive load \citep{8160794}.

\subsection{Mental Models of Dialog Agents}
With the increasing popularity of dialog agents, there has been a corresponding need to understand user's mental models, and how they influence interactions with such systems.
To date, the bulk of this research has focused on either single-turn, virtual assistants \citep{10.1145/3170427.3180286, 10.1145/2858036.2858288, zamora2017m, 9538817} or user interactions in cooperative games \cite{10.1145/3313831.3376316, weitz-etal-2021-fault, zhu2021understanding}. 
These studies have provided valuable insights into, e.g., the breadth of mental models users form about such systems \cite{10.1145/3170427.3180286} and how previous experience can shape user expectations and outcomes \citep{weitz-etal-2021-fault}. 
However, they represent only a small subset of types of dialog systems \citep{zhang2020recent, valizadeh-parde-2022-ai} and do not consider the role of adaptation.

\citet{10.1145/3411764.3445645} in contrast, chose to explore mental models around how students interacted with a virtual teaching assistant over a period of two months.
During this study, they found that it is possible to link linguistic information from students inputs with their self-reported mental models over the course of the interaction.
This finding supports our choice of adaptive agent, which infers a user's preferred interaction style from their input text.

To the best of our knowledge, \citet{10.1145/3290605.3300714} are the only study which explores user mental models around an adaptive dialog agent. 
They investigated a co-adapting agent, where the user explicitly tried to teach new behavior to a simulated personal assistant in a Wizard-of-Oz study. 
They found that users formed one of two main mental models of the agent, either viewing it on a technical or a social level, which 
greatly influenced how/what they attempted to teach it.
However, the mental models probed were intrinsically linked to the user's role as teacher.
We seek to expand on this research and explore user mental models in a more standard adaptive setting, where the dialog system implicitly updates its behavior without requiring users to take on an additional co-adapting role.

\section{Implementation}
\label{sec:implementation}
We choose to investigate the mental models of users in an information seeking setting, as this represents a common use case of modern dialog agents.
For the adaptive agent, we make use of the modified Conversational Tree Search (CTS) agent proposed by \citet{vath-etal-2024-cts-extension}, an open-source, adaptive dialog system for the domain of business travel.
For the non-adaptive baselines, we use a handcrafted dialog system and an FAQ system as they represent two of the most common interaction styles for information seeking systems.

\subsection{Adaptive Dialog System: CTS}

In the CTS task \citep{vath-etal-2023-conversational}, subject-area experts define a static dialog tree which encodes the possible conversation flows.
Nodes in the tree represent system utterances and edges the possible user responses.
The CTS agent is then trained on questions from users with different expectations, e.g, inputting a concrete question and getting a concrete answer or inputting a vague question and being asked follow-up questions/ given more background information until a concrete information need can be identified.
The agent uses reinforcement learning to adapt its behavior based on a user's input text, outputting or skipping over nodes in the graph as appropriate.

\begin{figure}[tb]
    \centering
    \includegraphics[width=0.4\textwidth]{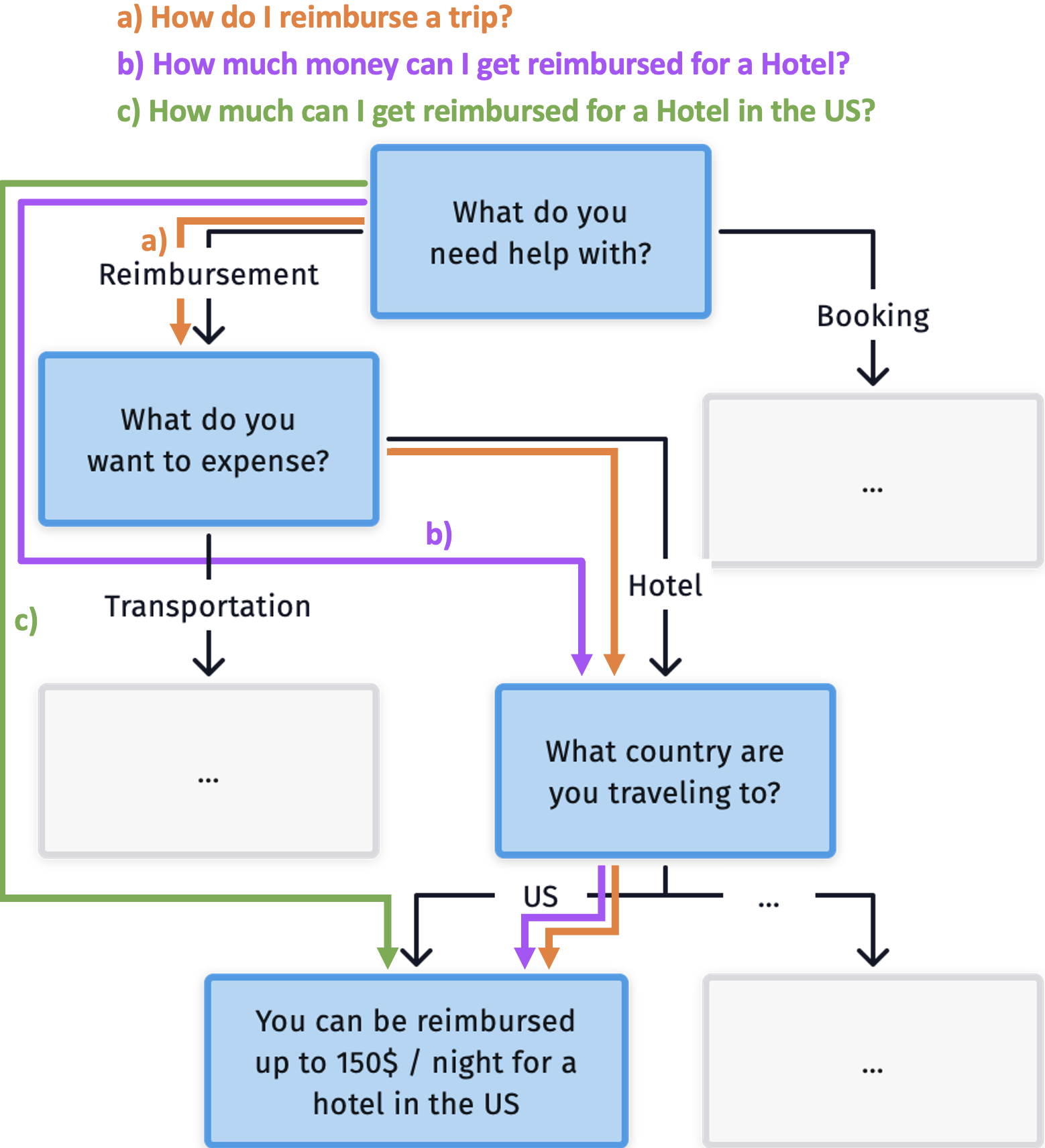}
    \caption{Illustration of the CTS agent behavior \cite{vath-etal-2023-conversational}. The agent adapts its behavior based on user expectations implicitly encoded in their input.}
    \label{fig:CTS_task}
\end{figure}

The extreme ends of the CTS agent's behavior can thus be modeled as a non-adaptive handcrafted dialog system (asking every node in the tree) or as an FAQ system (directly giving an answer), with the CTS agent able to adaptivly model the full spectrum of behavior between.
An illustration of this can be seen in Figure \ref{fig:CTS_task}, where different types of user inputs lead to different dialog system behavior. 

In this work, we follow the architecture and training process outlined in \cite{vath-etal-2024-cts-extension}; details in \autoref{sec:CTS_implementation}. All code is publicly available\footnote{GPL-3.0: \label{url}\url{www.removed-for-anonymity.github.com}}. 

\subsection{Baseline: Handcrafted Dialog System}
The handcrafted dialog agent (HDC) follows the dialog tree node-by-node, outputting node-text to the user at every step. 
If the user is asked a question, their response is then compared to the set of pre-defined answers associated with that node.
The agent then moves to the neighboring node connected by the answer with the highest similarity to the user's input (equivalent to ex. A in \autoref{fig:CTS_task}).
To compare results with the adaptive agent, we use the same dialog tree and the same state-of-the-art similarity model \cite{sentencetransformers} for encoding answers and user utterances.

\subsection{Baseline: FAQ System}
For the FAQ system, a user can input a specific question which will be matched against each of the nodes in the dialog graph.
The most similar node is then immediately returned to the user as the answer to their question (equivalent to ex. B in \autoref{fig:CTS_task}).
Again, we use the same dialog tree and similarity model.

\section{Dataset: RDMM}
The REIMBUSE Dialog Mental Models (\emph{RDMM}) dataset is a new dialog corpus, consisting of real user interaction logs with all three dialog systems (\autoref{sec:implementation}), surveys of each user's mental model of a chatbot before and after the interaction, and content analysis annotations (\autoref{sec:eval_methods}).
All three systems were trained on the REIMBURSE-En dataset \citep{vath-etal-2024-cts-extension}, for the domain of business travel reimbursement.
The dataset is publicly available\footnotemark[1].

For this corpus, we collect 188 dialogs from humans interacting with either the HDC baseline, the FAQ baseline, or the adaptive CTS agent.
Each dialog covers one of 14 different user information goals taken from the dialog graph, representing open-questions, general questions, and personalized questions (\autoref{sec:study_design}).
Below is an example dialog with the CTS agent:

\begin{dialogue}
\speak{GOAL} You want to know how the reimbursement process works for a research semester. You plan to bring your family with you.
\speak{USER} my family are coming with me, how do i get reimbursed for research semester
\speak{SYSTEM} How long will your research semester last?
\speak{USER} 1 month
\speak{SYSTEM} Your business trip must be planned as though you were traveling alone (without your family). Provide receipts [\textit{...}] 
\end{dialogue}

As the dialog interaction style between agents was quite different, dialogs with each system varied, e.g., in terms of dialog length and user utterance length. 
Dialog statistics are shown in \autoref{tab:corpus_stats}.

\begin{table}[tb]
    \center
    \resizebox{0.45\textwidth}{!}{
        \begin{tabular}{|l|c|c|c|}
            \hline
                                              & \textbf{FAQ}  & \textbf{HDC} & \textbf{Adaptive} \\
            \hline
             {\# Dialogs}                     & 61           & 66            & 61 \\
             {\# Successful dialogs}          & 35           & 29            & 47 \\
             {Avg. \# Turns/dialog}           & 2.3         & 13.3         & 7.4 \\ 
             {Avg. \# Words in Initial Input} & 10.2        & 8.2          & 8.7 \\
             {Avg. \# Words / Utterance }     & 10.2        & 5.4          & 6.4 \\
            \hline
        \end{tabular}
    }
    \caption{Corpus statistics for collected dialogs; numbers rounded to one decimal.}
    \label{tab:corpus_stats}
\end{table}

Each dialog in the corpus is also labeled with 1) the type of dialog system, 2) the information goal, 3) dialog length, 4) the end condition (success or failure), and 5) subjective user ratings for dialog length and quality of answer.

We further provide annotations of 1) each user's mental model of a dialog system before the interaction, 2) their mental model of the system after the interaction, and 3) their usability and trust ratings after the interaction.
Additional dialog examples can be seen in \autoref{sec:dialog_examples} and examples of mental model annotations can be seen in \autoref{tab:ContentAnalysis-Categories}.

\section{Study Design}
\label{sec:study_design}
We recruited participants from the United States, Great Britain, Australia, and Canada via the crowdsourcing platform Prolific \footnote{\url{https://prolific.com}}, paying at a rate of 9£/hr.
Interactions lasted on average 20 minutes.
Each participant was randomly assigned to interact with one of the three dialog systems.
We chose a between subjects design, as we did not want to influence users by exposing them to multiple dialog systems.

Participants were asked to complete a pre-survey, giving information about their domain familiarity and what mental models they had of dialog systems in general, before completing three different dialogs with their assigned system.
They were given no instruction on how they should interact with the dialog system.

The first dialog, each participant was randomly assigned an ``open'' goal. 
Open goals represented general topics in the dialog tree rather than any specific node and were intended to represent the information need of a user new to the domain.
For example: \emph{``you want to find out information on how to book a business trip.''} 
As we did not anticipate many users to be familiar with business travel regulations, we hoped to use these goals as a way to help familiarize participants with the domain.

For the second dialog, participants were randomly assigned an ``easy'' goal.
Easy goals represented an information need associated with a specific node in the graph, but did not require information about a user's specific case to be answered.
For example, \emph{``You want to know you can get reimbursed for a taxi''}.

For the final dialog, participants were randomly assigned to a ``hard'' goal.
Hard goals also represented information associated with a specific node in the graph, but required personalized details about the user's planned trip in order to answer.
\emph{``You want to know how much money you can get reimbursed for accommodation on your trip to France. You plan to stay with your brother''}.

Finally, users were asked to complete a post-survey with information about their mental model of the dialog system they interacted with and impressions of the interaction.

\section{Evaluation Methods}
\label{sec:eval_methods}
To understand their backgrounds, we asked participants to provide information about their age, gender, experience with dialog systems, and experience with business travel.
All surveys can be found in \autoref{sec:surveys}.

\paragraph{RQ1.1: Mental Models of Dialog Systems} As it is inherently difficult to measure mental models without also influencing them \citep{rowe1995measuring}, we take two complementary approaches. 

The first approach is a series of open-ended questions acting as a stand-in for think-aloud questions one would ask during a laboratory study \citep{friedman2018representing}.
We asked users about their expectations for both what type of input a task-oriented dialog system can understand and what type of answers it can generate,
e.g.,\textit{ ``How would you phrase your input to the chatbot? Is this similar or different to how you would use a search engine or ask a real person?''}.

The second approach was a series of Likert scale items asking users to rate how much they agreed with each statement. 
The first four statements related to their expectations for what type of input a dialog agent could understand and the second four to their expectations for what type of responses they could receive from a dialog system.
E.g., \textit{``In general I think that a chatbot can only give high-level/general answers to questions''}.

Each of the free response questions was analyzed using the standard content analysis technique  \citet{hsieh2005content_analysis}.
Following this technique, utterances are annotated with a fixed set of labels generated from the collected data, allowing trends to emerge from qualitative feedback.

\paragraph{RQ1.2: Effect of Mental Models on Interaction}
To assess what role mental models had on the interaction, we measured the objective length and success of each dialog.
Additionally, we asked users to rate the perceived length of the dialog on a scale from 1 (much too short) to 5 (much too long), as well as to rate their perception of the quality of how successful the dialog was on a scale from 1 (question not at all answered) to 4 (question completely answered).

\paragraph{RQ2.1: Effect of Adaptation on Mental Models}
Similar to how we measured mental models before the interaction, we asked users to both fill out free response and Likert items after the interaction.
However, in this case, free-response questions were based around the retrospective technique proposed by \citet{hoffman2018xaimetrics} and all questions were focused specifically on the interaction with the dialog system assigned to the user, rather than their perceptions of a generic dialog system.
Free response answers were again processed using content analysis.

\paragraph{RQ2.2: Effect of Adaptation on Success, Trust, and Usability}
To evaluate user trust, we use the reliability and trust subscales from the Trust in Automation (TiA) questionnaire \citep{korber2018theoretical}.
These subscales consist of six and two questions respectively, each rated on a five point Likert scale (1: strongly disagree to 5: strongly agree).

To measure perceived system usability, we use the four item Universal Measure of User Experience (UMUX) \citep{finstad2010usability}.
For measuring success, we use the same objective and subject measures as in RQ1.2.

\section{Pilot Study}
To validate our experimental design, we recruited 9 participants between the ages of 20 to 49.
Based on the feedback from this pilot study, we were able to fix technical errors in the study implementation and verify that the time needed by participants was in line with our estimates.

\section{Main Study}
For the main study, we recruited 66 participants from the USA, UK, Australia, and Canada.
Three participants were removed for not adhering to the task protocol or due to technical errors, resulting in a total of 63 participants across the three groups (CTS: 20, FAQ: 21, HDC: 22), and 188 dialogs.

Of the participants, 20 were male, 42 were female, and 1 person identified as other.
Their ages ranged from 20 to 69.
On average, participants had some familiarity with dialog systems (3 on a 5-point Likert item) and limited familiarity with business travel (1.9 on a 5-point Likert item).
There were no statistically significant differences in the distributions of gender, age, or previous experience between the three conditions. 

\section{Results \& Discussion}

In the following, we evaluate users as having a certain mental model if they rated that expectation with a value of 3 or higher in the pre-/post-survey.

\subsection{RQ1.1: Mental Models}
We explore user mental models both quantitatively and qualitatively.

\paragraph{Quantitative Analysis}

\begin{figure}
    \centering
    \includegraphics[width=0.49\textwidth]{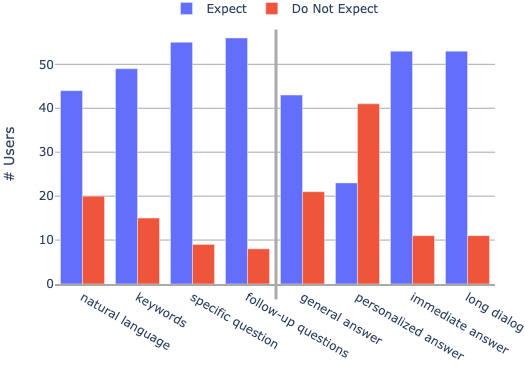}
    \caption{Distribution of mental models for what type of input a dialog system can understand (left 4 columns) and how it can respond (right 4 columns).}
    \label{fig:expecations}
\end{figure}

Looking at the user expectations in \autoref{fig:expecations}, we group user mental models into four categories about how they can interact with the system: expecting to 1) be able to give natural language input, 2) only give keyword input, 3) only be able to ask specific questions, and 4) be able to ask a general question which the system then clarifies. 
We also group expectations about system behavior into expecting 1) only general answers, 2) personalized answers to a user's exact situation, 3) direct answers (single-turn dialog), and 4) a long dialog with many questions before getting an answer.  
Here, we find there is no one expectation shared between all users, either for possible user inputs or system responses.
Even in cases where most users expect a certain behavior, there is still a sizable minority who do not.

\paragraph{Qualitative Analysis} 
To get a more granular understanding of users' expectations, we also performed qualitative analysis.
Most answers fell into one of five categories (see \autoref{tab:ContentAnalysis-Categories}): expectations about 1) the style of their input text, 2) the content of their input text 3) the style of the system's answer text 4) the content of the system's answer text, and 5) about the general dialog interaction.

\begin{table*}[htb]
    \caption{Main and sub-categories resulting from content analysis. For every sub-category (highlighted in bold), an example of participants' free-form feedback is given. Every example response is from a different participant.}
    \label{tab:ContentAnalysis-Categories}
    \resizebox{\linewidth}{!}{
    \centering
    \begin{tabular}{l|l}
    \hline
         A1 & \begin{tabular}[c]{@{}l@{}} I would phrase it much like asking a real person (\textbf{like a person)}\end{tabular} \\
         A2 & \begin{tabular}[c]{@{}l@{}} I would be more specific and direct with a chatbox than I would with a real person (\textbf{precise language})\end{tabular} \\
         A3 & \begin{tabular}[c]{@{}l@{}} I would use key terms, not necessarily full sentences. (\textbf{keywords})\end{tabular}\\
         A4 & \begin{tabular}[c]{@{}l@{}} I would use the chatbot like a search engine (\textbf{like a search engine})\end{tabular}\\
         \hline
         B1 &\begin{tabular}[c]{@{}l@{}} If the question is simple and commonly asked (\textbf{simple Questions})\end{tabular}\\
         \hline
         C1 & \begin{tabular}[c]{@{}l@{}} I would expect to get very general information from a chatbot that could be found on a companies website. (\textbf{generic information})\end{tabular} \\
         C2 & \begin{tabular}[c]{@{}l@{}} They should be able to provide prices, they should be able to provide the best options for you as a customer. (\textbf{Personalized})\end{tabular}\\
         C3 & \begin{tabular}[c]{@{}l@{}} and 100\% correct in the information I am being given.  (\textbf{correct/accurate})\end{tabular}\\ 
         C4 & \begin{tabular}[c]{@{}l@{}} Technically correct but perhaps not for the context - I would expect to check facts.  (\textbf{questionable accuracy})\end{tabular}\\ 
         \hline
         D1 & \begin{tabular}[c]{@{}l@{}} I would expect the level of information to be detailed and to a high level of knowledge (\textbf{highly detailed})\end{tabular} \\
         D2 & \begin{tabular}[c]{@{}l@{}} I would expect a clear, precise answer (\textbf{concise})\end{tabular}\\
         D3 & \begin{tabular}[c]{@{}l@{}} with sources I can verify myself if I doubt the accuracy of the information. (\textbf{with sources})\end{tabular}\\ 
         D4 & \begin{tabular}[c]{@{}l@{}} I would expect to get a similar style to what I might get from a real person nowadays. (\textbf{casual/friendly})\end{tabular}\\ 
         D5 & \begin{tabular}[c]{@{}l@{}} I would expect a formal answer
         (\textbf{stilted/formal})\end{tabular}\\ 
         \hline
         E1 & \begin{tabular}[c]{@{}l@{}} to save me time from having to browse through all the terms, rules and contracts myself. (\textbf{fast interaction})\end{tabular} \\
         E2 & \begin{tabular}[c]{@{}l@{}} and that it would be correct as long as I used the correct terms to ask. (\textbf{quality dependent on question}) \end{tabular} \\
         E3 & \begin{tabular}[c]{@{}l@{}} Unable to always answer my question. Sometimes go round in circles trying to get the information required (\textbf{unreliable})\end{tabular} \\
         \hline
         \multicolumn{2}{l}{ \textbf{A} = User input style, \hspace{0.5cm}\textbf{B} = User input content, \hspace{0.5cm}\textbf{C} = Dialog agent output content, \hspace{0.5cm} \textbf{D} = Dialog agent output style, \hspace{0.5cm} \textbf{E} = Interaction}\\
    \end{tabular}
    }
\end{table*}
We find that users have very different, and in many cases contradictory, expectations for how they should interact with a dialog agent, and how the agent, in turn, should behave.
The contradictory nature of these mental models further supports the need for reliable adaptive dialog systems.

\subsection{RQ1.2: Effect of Mental Models on Interaction}

To understand how these mental models translate to actual user interactions,
we look for any statistically significant differences between users who share a (quantitative) mental model and those who do not in terms of objective measures, i.e., dialog length and success, and subjective measures, i.e., usability, reliability, and trust.
\begin{table}[htb]
    \center
    \resizebox{0.49\textwidth}{!}{
         \begin{tabular}{ |c|c|c|c|c|c| } 
         \hline
          \textbf{Mental models} & 
          \makecell{\textbf{Length}}  & 
          \makecell{\textbf{Success}} & 
          \makecell{\textbf{Usability}} &
          \textbf{Reliability} &
          \textbf{Trust}
          \\ \hline

         Natural language            &           &      &               & CTS          & CTS  \\
         Only Keywords               &           &      & CTS, HDC      & HDC          &    \\
         Only Specific Questions     &  CTS      & CTS  & CTS, HDC      &              &    \\
         Vague Questions             &           & HDC  & HDC           & HDC          & HDC  \\
         \hline
         Only General Answers        &   CTS     &      &               &               &   \\
         Personalized Answers        & CTS, HDC  & FAQ  &               &               &   \\
         Single-Turn Interaction     &           &      &  FAQ          &               &   \\
         Multi-Turn Interaction      &           & FAQ  & CTS, HDC, FAQ & CTS, HDC, FAQ & CTS, FAQ  \\
         \hline
        \end{tabular}
    }
    \caption{Significant effects ($p<0.05$) of mental models on objective and subjective metrics per dialog system.}
    \label{tab:influence_of_mms}
\end{table}

Using standard t-tests, we find that all 8 expectations had a statistically significant effect ($p<0.05$) on either objective or subjective evaluation metrics. 
These effects depended on the dialog system users interacted with (\autoref{tab:influence_of_mms}).
For example, user expecting a multi-turn interaction or personalized answers were less successful when interacting with an FAQ system than those who did not.
Those expecting multi-turn dialogs additionally found the FAQ system significantly less usable.
Conversely, users of the HDC system who expected that they could phrase general questions and the dialog system would help narrow down their information need were more successful on all of these same metrics than those who expected to need a specific question.
For the adaptive and HDC dialog systems, the expectation that the dialog system would only recognize keywords and the expectation that users needed to be able to phrase a specific question also influenced success and usability.

In general, we found the greatest impacts of mental models in cases where a mismatch existed between user expectations and dialog system behavior.
These results are line with work by \citet{10.1145/2207676.2207678} and \citet{bansal2019beyond}.

\subsection{RQ2.1: Effect of Adaptation on Mental Models}

\begin{table}[htb]
    \center
    \resizebox{0.48\textwidth}{!}{
         \begin{tabular}{ |c|c|c|c| } 
         \hline
          \textbf{Mental models} & 
          \makecell{\textbf{Adaptive}}  & 
          \makecell{\textbf{FAQ}} & 
          \makecell{\textbf{HDC} } 
          \\ \hline

         Natural language             & 0.10        &  0.02     & -0.64   \\
         Only Keywords                & 0.20        & -0.05     &  0.23    \\
         Only Specific Questions       & \textbf{-0.75**}     & -0.29     &  0.09   \\
         Vague Questions + Clarification  & 0.35    & \textbf{-2.35***}  & -0.64   \\
         \hline
         Only General Answers         & 0.10        &  \textbf{1.15**}   &  0.27   \\
         Personalized Answers         & \textbf{0.65*}       & -0.57     & -0.14   \\
         Single-Turn Interaction      & -0.30       &  0.38     & -0.55   \\
         Multi-Turn Interaction       & 0.28        & \textbf{-1.52**}   &  0.27   \\
         \hline
        \end{tabular}
    }
    \caption{Change in mental models before and after the interaction for each system. *$p<0.05$; **$p<0.01$; ***$p<0.001$.}
    \label{tab:updated_models}
\end{table}

When looking at changes in users' mental models after interacting with the adaptive system (see \autoref{tab:updated_models}), we find that users' models for the adaptive agent remain largely constant, except for two cases where the agent exceeded their expectations. 
This suggests that the adaptive system was able to shape its behavior to match the users' expected interaction styles.
In contrast, FAQ users had to update models related both to interaction and to system answers after interacting with their dialog systems.
While users from the HDC group also had no significant changes to their mental models, this seems to be due to negative expectations being met.
Looking at the qualitative feedback, for example, 8 out of the 22 participants who interacted with the HDC system did not think that the dialog system was able to understand their input, compared to 0 participants from the adaptive system group.

From this, we see that the adaptive agent is able to implicitly adapt in a way that remains in line with user expectations.
Furthermore, our results show that through this adaptation, it is possible for an adaptive agent to overcome users' negative preconceptions.

\subsection{RQ2.2: Effect of Adaptation on Usability, Trust, \& Task Success}

We evaluate both the objective success and length of the dialog and the users' subjective perception of these metrics.
Additionally, we look at the effects adaptation has on the quality of each dialog, comparing user ratings for usability, reliability, and trust for each type of dialog system. 

\label{sec:human_obj}
\begin{table}[htb]
    \center
    \resizebox{0.49\textwidth}{!}{
         \begin{tabular}{ |c|c|c|c|c| } 
         \hline
          \textbf{Model} & 
          \makecell{\textbf{\# Turns}}  & 
          \makecell{\textbf{Success}} & 
          \makecell{\textbf{Perceived} \\ \textbf{Length}} & 
          \makecell{\textbf{Answer} \\ \textbf{Satisfaction}} 
          \\ \hline

         Adaptive  & 7.38  & \textbf{77.05}  & 2.92 &  \textbf{2.87}  \\
         FAQ  & \textbf{2.26}  & 57.38  & 2.28 &  2.61  \\
         HDC  & 13.32 & 43.94  & \textbf{3.08} &  2.41  \\
         \hline
        \end{tabular}
    }
    \caption{Average objective and subjective performance metrics per dialog system. Perceived length is measured on a 5-point scale, perceived quality on a 4-point scale.}
    \label{tab:adaptation_success}
\end{table}

\begin{table}[htb]
    \center
    \resizebox{0.375\textwidth}{!}{
         \begin{tabular}{ |c|c|c|c| } 
         \hline
          \textbf{Model} & 
          \makecell{\textbf{Trust}}  & 
          \makecell{\textbf{Reliability}} & 
          \makecell{\textbf{Usability} } 
          \\ \hline

         Adaptive  & \textbf{3.16}  & \textbf{2.96}  & \textbf{62.83}   \\
         FAQ  & 2.83  & 2.79  & 57.73   \\
         HDC  & 2.61 & 2.42  & 36.93   \\
         \hline
        \end{tabular}
    }
    \caption{Averaged trust, reliability, and usability ratings. The adaptive agent and the FAQ system were significantly more usable ($p < 0.05$) than the HDC system.}
    \label{tab:adaptation_trust}
\end{table}

On objective measures (\autoref{tab:adaptation_success}), the adaptive agent is significantly more successful ($77.05\%$) than either the FAQ ($57.38\%$) or HDC ($43.94\%$) system ($p<0.05$ and $p<0.001$ respectively; Tukey test), while at the same time, requiring significantly ($p<0.001$) fewer turns (7.38 turns) than the HDC policy (13.32 turns).
Compared to the FAQ system, where dialogs were rated as too short (2.28 out of 5), the adaptive system (2.92 out of 5) was subjectively perceived to have a more appropriate dialog length (Tukey test; $p<0.001$).
While these differences hold for all types of goals, the benefit of adaptation is most apparent for dialogs which required a personalized answer.
Here, the adaptive agent leads to successful dialogs $57.89\%$ of the time, FAQ system $25.00\%$, and HDC system $22.73\%$.
All models struggle with these harder goals, but the adaptive agent performs significantly (Tukey test; $p< 0.05$) better.

When taken together, these results suggest that adapting to the user's interaction style has a positive effect, not only increasing task success (especially in more complex cases), but also creating a comfortable middle-ground between too short single-turn FAQ and too long handcrafted dialogs.

When looking at the user trust, reliability, and usability ratings in \autoref{tab:adaptation_trust}, we see that both the FAQ and adaptive systems were rated significantly more usable ($p<0.05$ Tukey test) than the handcrafted dialog policy.
This supports the results from Section \ref{sec:human_obj}, and suggests that the burden of answering follow-up questions in the adaptive agent -- and thus extending the dialog length-- is no higher for users than that of the single-turn dialog scenario in the FAQ setting.
Although the adaptive agent scores higher in trust and reliability, the effects were not found to be significant, perhaps in part due to the users' relative unfamiliarity with the domain and the corresponding difficulty of determining if a given response was correct or not.
However, in contrast to previous work \citep{bansal2019beyond}, it does suggest that implicit adaptation does not necessarily degrade human-AI performance.

\section{Conclusion}
In this work, we investigate what mental models users have about information seeking dialog systems, how these mental models impact users' interactions with (adaptive) dialog systems, how these mental models are updated through such interaction, and how adaptation affects user trust and usability ratings. 
We additionally provide a new corpus of dialogs, self-reported mental models, and content analysis annotations, which can be used for studying mental models and their implications for interaction with task-oriented dialog agents.

We find that users form a variety of mental models about how they can interact with a dialog agent and that these mental models were often contradictory between users, e.g, that input should be as specific as possible vs. general questions were better.
We further find that users had varying expectations for how the system could answer, e.g., asking follow-up questions or directly returning an answer.
These mental models significantly impacted how users engaged with the dialog system and how successful they were at finding the information they were looking for, highlighting the importance of aligning dialog system functionality with user expectations.

Furthermore, we demonstrate that the implicit adaptation carried out by the dialog agent largely did not update users' mental models, with the exception of even improving two negative expectations users had about limitations of a dialog system.
Rather, the adaptive agent was able to implicitly update its behavior to align better with the users' mental models.
Adapting in this way significantly improved objective evaluation metrics, i.e. dialog success and dialog efficiency, as well as the subjective metrics: usability, perceived length, and perceived answer quality.

In summary, while \citep{bansal2019beyond} found implicit adaptation to be harmful, we conclude that implicit adaptation in line with user expectations can significantly improve both the success and usability of a dialog agent.
Additionally, in contrast to the results of \citet{10.1145/3290605.3300714}, we find that a teaching period may not be required before proactive adaptation can take place, so long as that adaptation is aligned with users' expectations.
This, however, only serves to underline the importance of first understanding user mental models either through direct evaluation or implicitly deriving this information from user input.

\section{Limitations}
While we tried to recruit a diverse background of participants, the study was conducted in English with participants living in English-speaking countries, which may bias the results.
Additionally, we explore only one possible axis of adaptation, namely how many follow-up questions/ how much additional information should be asked or skipped before delivering users an answer.
However, during our analysis of users' mental models before the interaction, we found that there are multiple axes in which users have contradicting expectations, e.g., level of answer detail, linguistic style of dialog system output, etc.
In the future it would be interesting to explore how each of these axes affect the dialog interaction, either individually or in combination.

\section{Ethical Considerations and Risks}
Before starting the study, we performed power analysis to ensure that we would have enough power to detect an effect if one were present, so as to not waste the time of any participants who took part in the study.
To ensure that users could give informed consent, we provided a detailed description of the task and research objectives both on the crowdsourcing platform and once they had accepted the task.
In respect of participant privacy, we specifically did not collect personally identifying data from any users.
To this end, we store all logs and survey responses using an anonymous hash generated based on a given username, rather than with the username itself.
In this way, users could log in again if they needed to take a break in the middle of the interaction, but we had no way of directly linking any recorded results to, e.g., users' Prolific account identifiers.
To ensure that participants were fairly compensated, we followed best practices recommended by the crowdsourcing platform paying users at 9£/hr, which was in-line with minimum wage in the countries we recruited participants from at the time of the experiment.
We additionally used our pilot study to verify that our estimated time was below the median time we selected when advertising the task, meaning most participants had a higher hourly wage.

In terms of risks, the goal of this paper is to lay a ground-work for creating more effective adaptive dialog agents.
However, this does have the possible risk of creating chatbots which could also be used to more effectively replace human jobs.

\bibliography{custom}

\begin{thebibliography}{40}
\providecommand{\natexlab}[1]{#1}

\bibitem[{Bansal et~al.(2019{\natexlab{a}})Bansal, Nushi, Kamar, Lasecki, Weld,
  and Horvitz}]{bansal2019beyond}
Gagan Bansal, Besmira Nushi, Ece Kamar, Walter~S Lasecki, Daniel~S Weld, and
  Eric Horvitz. 2019{\natexlab{a}}.
\newblock Beyond accuracy: The role of mental models in human-ai team
  performance.
\newblock In \emph{Proceedings of the AAAI conference on human computation and
  crowdsourcing}, volume~7, pages 2--11.

\bibitem[{Bansal et~al.(2019{\natexlab{b}})Bansal, Nushi, Kamar, Weld, Lasecki,
  and Horvitz}]{bansal2019updates}
Gagan Bansal, Besmira Nushi, Ece Kamar, Daniel~S Weld, Walter~S Lasecki, and
  Eric Horvitz. 2019{\natexlab{b}}.
\newblock Updates in human-ai teams: Understanding and addressing the
  performance/compatibility tradeoff.
\newblock In \emph{Proceedings of the AAAI Conference on Artificial
  Intelligence}, volume~33, pages 2429--2437.

\bibitem[{Brachman et~al.(2023)Brachman, Pan, Do, Dugan, Chaudhary, Johnson,
  Rai, Chakraborti, Gschwind, Laredo et~al.}]{brachman2023follow}
Michelle Brachman, Qian Pan, Hyo~Jin Do, Casey Dugan, Arunima Chaudhary,
  James~M Johnson, Priyanshu Rai, Tathagata Chakraborti, Thomas Gschwind, Jim~A
  Laredo, et~al. 2023.
\newblock Follow the successful herd: Towards explanations for improved use and
  mental models of natural language systems.
\newblock In \emph{Proceedings of the 28th International Conference on
  Intelligent User Interfaces}, pages 220--239.

\bibitem[{Chen and Pu(2012)}]{chen2012critiquing}
Li~Chen and Pearl Pu. 2012.
\newblock Critiquing-based recommenders: survey and emerging trends.
\newblock \emph{User Modeling and User-Adapted Interaction}, 22(1):125--150.

\bibitem[{Cho(2018)}]{10.1145/3170427.3180286}
Janghee Cho. 2018.
\newblock \href {https://doi.org/10.1145/3170427.3180286} {Mental models and
  home virtual assistants (hvas)}.
\newblock In \emph{Extended Abstracts of the 2018 CHI Conference on Human
  Factors in Computing Systems}, CHI EA '18, page 1–6, New York, NY, USA.
  Association for Computing Machinery.

\bibitem[{Clements(2004)}]{clements2004perspective}
DH~Clements. 2004.
\newblock Perspective on “the child’s thought and geometry”.
\newblock \emph{Classics in mathematics education research}, pages 60--66.

\bibitem[{Finstad(2010)}]{finstad2010usability}
Kraig Finstad. 2010.
\newblock The usability metric for user experience.
\newblock \emph{Interacting with computers}, 22(5):323--327.

\bibitem[{Firdaus et~al.(2023)Firdaus, Shandilya, Ekbal, and
  Bhattacharyya}]{9801557}
Mauajama Firdaus, Arunav Shandilya, Asif Ekbal, and Pushpak Bhattacharyya.
  2023.
\newblock \href {https://doi.org/10.1109/TCSS.2022.3182986} {Being polite:
  Modeling politeness variation in a personalized dialog agent}.
\newblock \emph{IEEE Transactions on Computational Social Systems},
  10(4):1455--1464.

\bibitem[{Friedman et~al.(2018)Friedman, Forbus, and
  Sherin}]{friedman2018representing}
Scott Friedman, Kenneth Forbus, and Bruce Sherin. 2018.
\newblock Representing, running, and revising mental models: A computational
  model.
\newblock \emph{Cognitive Science}, 42(4):1110--1145.

\bibitem[{Gero et~al.(2020)Gero, Ashktorab, Dugan, Pan, Johnson, Geyer, Ruiz,
  Miller, Millen, Campbell, Kumaravel, and Zhang}]{10.1145/3313831.3376316}
Katy~Ilonka Gero, Zahra Ashktorab, Casey Dugan, Qian Pan, James Johnson, Werner
  Geyer, Maria Ruiz, Sarah Miller, David~R. Millen, Murray Campbell, Sadhana
  Kumaravel, and Wei Zhang. 2020.
\newblock \href {https://doi.org/10.1145/3313831.3376316} {Mental models of ai
  agents in a cooperative game setting}.
\newblock In \emph{Proceedings of the 2020 CHI Conference on Human Factors in
  Computing Systems}, CHI '20, page 1–12, New York, NY, USA. Association for
  Computing Machinery.

\bibitem[{Halasz and Moran(1983)}]{halasz1983mental}
Frank~G Halasz and Thomas~P Moran. 1983.
\newblock \href {https://doi.org/10.1145/800045.801613} {Mental models and
  problem solving in using a calculator}.
\newblock In \emph{Proceedings of the SIGCHI conference on Human Factors in
  Computing Systems}, pages 212--216.

\bibitem[{He et~al.(2023)He, Wang, Ding, and Shen}]{he2023conversation}
Ming He, Jiwen Wang, Tianyu Ding, and Tong Shen. 2023.
\newblock Conversation and recommendation: knowledge-enhanced personalized
  dialog system.
\newblock \emph{Knowledge and Information Systems}, 65(1):261--279.

\bibitem[{Hoffman et~al.(2018)Hoffman, Mueller, Klein, and
  Litman}]{hoffman2018xaimetrics}
Robert~R. Hoffman, Shane~T. Mueller, Gary Klein, and Jordan Litman. 2018.
\newblock \href {http://arxiv.org/abs/1812.04608} {Metrics for explainable
  {AI:} challenges and prospects}.
\newblock \emph{CoRR}, abs/1812.04608.

\bibitem[{Hsieh and Shannon(2005)}]{hsieh2005content_analysis}
Hsiu-Fang Hsieh and Sarah~E Shannon. 2005.
\newblock Three approaches to qualitative content analysis.
\newblock \emph{Qualitative health research}, 15(9):1277--1288.

\bibitem[{Höök(2000)}]{8160794}
K.~Höök. 2000.
\newblock \href {https://doi.org/10.1016/S0953-5438(99)00006-5} {Steps to take
  before intelligent user interfaces become real}.
\newblock \emph{Interacting with Computers}, 12(4):409--426.

\bibitem[{Janarthanam and Lemon(2014)}]{janarthanam-lemon-2014-adaptive}
Srinivasan Janarthanam and Oliver Lemon. 2014.
\newblock \href {https://doi.org/10.1162/COLI_a_00203} {Adaptive generation in
  dialogue systems using dynamic user modeling}.
\newblock \emph{Computational Linguistics}, 40(4):883--920.

\bibitem[{Johnson-Laird(1980)}]{johnson1980mental}
Philip~N Johnson-Laird. 1980.
\newblock Mental models in cognitive science.
\newblock \emph{Cognitive science}, 4(1):71--115.

\bibitem[{Kim and Lim(2019)}]{10.1145/3290605.3300714}
Da-jung Kim and Youn-kyung Lim. 2019.
\newblock \href {https://doi.org/10.1145/3290605.3300714} {Co-performing agent:
  Design for building user-agent partnership in learning and adaptive
  services}.
\newblock In \emph{Proceedings of the 2019 CHI Conference on Human Factors in
  Computing Systems}, CHI '19, page 1–14, New York, NY, USA. Association for
  Computing Machinery.

\bibitem[{K{\"o}rber(2018)}]{korber2018theoretical}
Moritz K{\"o}rber. 2018.
\newblock Theoretical considerations and development of a questionnaire to
  measure trust in automation.
\newblock In \emph{Congress of the International Ergonomics Association}, pages
  13--30. Springer.

\bibitem[{Kulesza et~al.(2012)Kulesza, Stumpf, Burnett, and
  Kwan}]{10.1145/2207676.2207678}
Todd Kulesza, Simone Stumpf, Margaret Burnett, and Irwin Kwan. 2012.
\newblock \href {https://doi.org/10.1145/2207676.2207678} {Tell me more? the
  effects of mental model soundness on personalizing an intelligent agent}.
\newblock In \emph{Proceedings of the SIGCHI Conference on Human Factors in
  Computing Systems}, CHI '12, page 1–10, New York, NY, USA. Association for
  Computing Machinery.

\bibitem[{Luger and Sellen(2016)}]{10.1145/2858036.2858288}
Ewa Luger and Abigail Sellen. 2016.
\newblock \href {https://doi.org/10.1145/2858036.2858288} {"like having a
  really bad pa": The gulf between user expectation and experience of
  conversational agents}.
\newblock In \emph{Proceedings of the 2016 CHI Conference on Human Factors in
  Computing Systems}, CHI '16, page 5286–5297, New York, NY, USA. Association
  for Computing Machinery.

\bibitem[{Ma et~al.(2020)Ma, Nguyen, Xing, and Cambria}]{MA202050}
Yukun Ma, Khanh~Linh Nguyen, Frank~Z. Xing, and Erik Cambria. 2020.
\newblock \href {https://doi.org/10.1016/j.inffus.2020.06.011} {A survey on
  empathetic dialogue systems}.
\newblock \emph{Information Fusion}, 64:50--70.

\bibitem[{Narducci et~al.(2018)Narducci, de~Gemmis, Lops, and
  Semeraro}]{narducci2018improving}
Fedelucio Narducci, Marco de~Gemmis, Pasquale Lops, and Giovanni Semeraro.
  2018.
\newblock Improving the user experience with a conversational recommender
  system.
\newblock In \emph{International Conference of the Italian Association for
  Artificial Intelligence}, pages 528--538. Springer.

\bibitem[{Norman(2014)}]{norman2014some}
Donald~A Norman. 2014.
\newblock \href {https://doi.org/10.4324/978131580272} {Some observations on
  mental models}.
\newblock In \emph{Mental models}, pages 15--22. Psychology Press.

\bibitem[{Reimers and Gurevych(2019)}]{sentencetransformers}
Nils Reimers and Iryna Gurevych. 2019.
\newblock \href {https://doi.org/10.18653/v1/D19-1410} {Sentence-bert: Sentence
  embeddings using siamese bert-networks}.
\newblock In \emph{Proceedings of the 2019 Conference on Empirical Methods in
  Natural Language Processing and the 9th International Joint Conference on
  Natural Language Processing, {EMNLP-IJCNLP} 2019, Hong Kong, China, November
  3-7, 2019}, pages 3980--3990. Association for Computational Linguistics.

\bibitem[{Ritschel and Andr{\'e}(2018)}]{ritschel2018shaping}
Hannes Ritschel and Elisabeth Andr{\'e}. 2018.
\newblock Shaping a social robot’s humor with natural language generation and
  socially-aware reinforcement learning.
\newblock In \emph{Proceedings of the workshop on NLG for human--robot
  interaction}, pages 12--16.

\bibitem[{Rowe and Cooke(1995)}]{rowe1995measuring}
Anna~L Rowe and Nancy~J Cooke. 1995.
\newblock Measuring mental models: Choosing the right tools for the job.
\newblock \emph{Human resource development quarterly}, 6(3):243--255.

\bibitem[{Rutjes et~al.(2019)Rutjes, Willemsen, and
  IJsselsteijn}]{Rutjes2019AIHCI}
Heleen Rutjes, Martijn Willemsen, and Wijnand IJsselsteijn. 2019.
\newblock Considerations on explainable ai and users’ mental models.
\newblock In \emph{Where is the Human? Bridging the Gap Between AI and HCI},
  United States. Association for Computing Machinery, Inc.

\bibitem[{Tenhundfeld et~al.(2022)Tenhundfeld, Barr, O'Hear, and
  Weger}]{9538817}
Nathan~L. Tenhundfeld, Hannah~M. Barr, Emily~H. O'Hear, and Kristin Weger.
  2022.
\newblock \href {https://doi.org/10.1109/THMS.2021.3107493} {Is my siri the
  same as your siri? an exploration of users’ mental model of virtual
  personal assistants, implications for trust}.
\newblock \emph{IEEE Transactions on Human-Machine Systems}, 52(3):512--521.

\bibitem[{Valizadeh and Parde(2022)}]{valizadeh-parde-2022-ai}
Mina Valizadeh and Natalie Parde. 2022.
\newblock \href {https://doi.org/10.18653/v1/2022.acl-long.458} {The {AI}
  doctor is in: A survey of task-oriented dialogue systems for healthcare
  applications}.
\newblock In \emph{Proceedings of the 60th Annual Meeting of the Association
  for Computational Linguistics (Volume 1: Long Papers)}, pages 6638--6660,
  Dublin, Ireland. Association for Computational Linguistics.

\bibitem[{V{\"a}th et~al.(2023)V{\"a}th, Vanderlyn, and
  Vu}]{vath-etal-2023-conversational}
Dirk V{\"a}th, Lindsey Vanderlyn, and Ngoc~Thang Vu. 2023.
\newblock \href {https://doi.org/10.18653/v1/2023.eacl-main.91} {Conversational
  tree search: A new hybrid dialog task}.
\newblock In \emph{Proceedings of the 17th Conference of the European Chapter
  of the Association for Computational Linguistics}, pages 1264--1280,
  Dubrovnik, Croatia. Association for Computational Linguistics.

\bibitem[{V{\"a}th et~al.(2024)V{\"a}th, Vanderlyn, and
  Vu}]{vath-etal-2024-cts-extension}
Dirk V{\"a}th, Lindsey Vanderlyn, and Ngoc~Thang Vu. 2024.
\newblock \href {https://aclanthology.org/2024.lrec-main.1428} {Towards a
  zero-data, controllable, adaptive dialog system}.
\newblock In \emph{Proceedings of the 2024 Joint International Conference on
  Computational Linguistics, Language Resources and Evaluation (LREC-COLING
  2024)}, pages 16433--16449, Torino, Italia. ELRA and ICCL.

\bibitem[{Wang et~al.(2021)Wang, Saha, Gregori, Joyner, and
  Goel}]{10.1145/3411764.3445645}
Qiaosi Wang, Koustuv Saha, Eric Gregori, David Joyner, and Ashok Goel. 2021.
\newblock \href {https://doi.org/10.1145/3411764.3445645} {Towards mutual
  theory of mind in human-ai interaction: How language reflects what students
  perceive about a virtual teaching assistant}.
\newblock In \emph{Proceedings of the 2021 CHI Conference on Human Factors in
  Computing Systems}, CHI '21, New York, NY, USA. Association for Computing
  Machinery.

\bibitem[{Weitz et~al.(2021)Weitz, Vanderlyn, Vu, and
  Andr{\'e}}]{weitz-etal-2021-fault}
Katharina Weitz, Lindsey Vanderlyn, Ngoc~Thang Vu, and Elisabeth Andr{\'e}.
  2021.
\newblock \href {https://doi.org/10.18653/v1/2021.conll-1.1} {{``}it{'}s our
  fault!{''}: Insights into users{'} understanding and interaction with an
  explanatory collaborative dialog system}.
\newblock In \emph{Proceedings of the 25th Conference on Computational Natural
  Language Learning}, pages 1--16, Online. Association for Computational
  Linguistics.

\bibitem[{Weld et~al.(2003)Weld, Anderson, Domingos, Etzioni, Gajos, Lau, and
  Wolfman}]{weld2003automatically}
Daniel Weld, Corin Anderson, Pedro Domingos, Oren Etzioni, Krzysztof~Z Gajos,
  Tessa Lau, and Steve Wolfman. 2003.
\newblock Automatically personalizing user interfaces.

\bibitem[{Yang et~al.(2018)Yang, Qu, Lei, Zhu, Zhao, Chen, and
  Huang}]{doi:10.1137/1.9781611975321.71}
Min Yang, Qiang Qu, Kai Lei, Jia Zhu, Zhou Zhao, Xiaojun Chen, and Joshua~Z.
  Huang. 2018.
\newblock \href {https://doi.org/10.1137/1.9781611975321.71}
  {\emph{Investigating Deep Reinforcement Learning Techniques in Personalized
  Dialogue Generation}}, pages 630--638.

\bibitem[{Zamora(2017)}]{zamora2017m}
Jennifer Zamora. 2017.
\newblock I'm sorry, dave, i'm afraid i can't do that: Chatbot perception and
  expectations.
\newblock In \emph{Proceedings of the 5th international conference on human
  agent interaction}, pages 253--260.

\bibitem[{Zhang et~al.(2020)Zhang, Takanobu, Zhu, Huang, and
  Zhu}]{zhang2020recent}
Zheng Zhang, Ryuichi Takanobu, Qi~Zhu, MinLie Huang, and XiaoYan Zhu. 2020.
\newblock \href {https://doi.org/10.1007/s11431-020-1692-3} {Recent advances
  and challenges in task-oriented dialog systems}.
\newblock \emph{Science China Technological Sciences}, 63(10):2011--2027.

\bibitem[{Zhu and Villareale(2021)}]{zhu2021understanding}
Jichen Zhu and Jennifer Villareale. 2021.
\newblock Understanding mental models of ai through player-ai interaction.
\newblock In \emph{Extended Abstracts of the 2021 CHI Conference on Human
  Factors in Computing Systems (CHI EA'21)}, page~11.

\bibitem[{Zimmerman et~al.(2007)Zimmerman, Tomasic, Simmons, Hargraves,
  Mohnkern, Cornwell, and McGuire}]{10.1145/1240624.1240843}
John Zimmerman, Anthony Tomasic, Isaac Simmons, Ian Hargraves, Ken Mohnkern,
  Jason Cornwell, and Robert~Martin McGuire. 2007.
\newblock \href {https://doi.org/10.1145/1240624.1240843} {Vio: A
  mixed-initiative approach to learning and automating procedural update
  tasks}.
\newblock In \emph{Proceedings of the SIGCHI Conference on Human Factors in
  Computing Systems}, CHI '07, page 1445–1454, New York, NY, USA. Association
  for Computing Machinery.

\end{thebibliography}

\appendix
\onecolumn

\section{Adaptive Dialog Agent Implementation}
\label{sec:CTS_implementation}
The CTS agent was published under the GPL-3.0 license, making our use of it as the basis for our adaptive agent consistent with its intended use.

\subsection{Conversational Tree Search Task}
Given a dialog tree (e.g. \autoref{fig:CTS_task}), the goal of the CTS task is to efficiently traverse this tree in order to answer a user's information need \citep{vath-etal-2023-conversational}.
A Reinforcement Learning policy is trained to either output text at the current node (e.g., asking a question or giving information), or to skip that node and directly move to a neighbouring node.. 

In order to model different styles of user interaction (general domain exploration vs specific questions), there are two goal settings within this task framework:
\begin{itemize}
    \item \textbf{Guided Dialog}: This models scenarios where a user has a vague information goal. Rather than posing a concrete question, the user needs to be guided through the dialog graph, exploring the domain.

    \item \textbf{Free Dialog}: Free dialog, in contrast, models users who have a concrete information need and expect the system to be able to answer it as directly as possible. If the system is not sure about an upcoming decision, it may choose to ask follow-up questions to increase its understanding of the user's goal. Each turn serves to clarify the goal or skip closer to the answer.
\end{itemize}

\subsection{Evaluation Objectives}
The objective dialog metrics and the rewards for the RL agent were taken from the modified evaluation method in \cite{vath-etal-2024-cts-extension}, which draws a concrete goal for users in both types of dialog instead of the original method \citep{vath-etal-2023-conversational}, which only considered turn-wise goals (agent only needs to navigate to the correct follow-up node) for guided-mode.
In short, the evaluation objectives used in this paper are:
\begin{itemize}
     \item \textbf{Free Mode}: In free mode, the objective is to maximize both \textit{task success} (reaching a final, pre-drawn goal node) and the \textit{skip ratio} (percentage of nodes in the dialog which are skipped instead of outputted to the user).
    \item \textbf{Guided Dialog}: For guided dialog, the objective is to maximize \textit{task success} while minimizing the \textit{skip ratio}.
\end{itemize}

\subsection{RL Model Parameters and Training Resources}
\begin{table}[h!]
    \centering
    \begin{tabular}{c|c}
        \textbf{Parameter} & \textbf{Value} \\ \hline
        Layer type & Linear \\
        Activation (after each layer except in Dialog Mode Classifier Head) & SELU \\
        Shared Layer Neurons (one value / layer) & $8096, 4096, 4096$ \\
        Value Function Layer Neurons (one value / layer) & $2048, 1024$ \\
        Advantage Function Layer Neurons (one value / layer) & $4096, 2048, 1024$ \\
        Dialog Mode Classifier Neurons (one value / layer) & $256, 1$ \\
        Dropout (after each layer) & $25$\% 
    \end{tabular}
\end{table}

The agent was trained on a single RTX 3090 GPU. 
In total, we required approximately 840 total hours including parameter tuning and training.

\newpage
\subsection{RL Training Parameters}

The following parameters were used to train the CTS agent (chosen through manual tuning) with performance measured against a user simulator:
\begin{table}[h]
    \centering
    \begin{tabular}{c|c}
         \textbf{Parameter} & \textbf{Value} \\ \hline
         Optimizer & Adam \\
         Learning Rate & $1e^{-4}$ \\
         $\lambda$ & $0.1$ \\
         Maximum Training Dialog Turns & $2M$ \\
         Max. Gradient Norm & $1.0$ \\
         Batch Size & $256$ \\
         $\gamma$ & $0.99$ \\
         Exploration fraction of Training Turns & $0.99$ \\
         Exploration Scheme & $\epsilon$-greedy \\
         $\epsilon$ start & $0.6$ \\
         $\epsilon$ end & $0.0$ \\
         Training frequency (w.r.t. dialog turns) & $3$ \\
         Training start (w.r.t. dialog turns) & $1280$ \\
         DDQN Target Network update frequency (w.r.t. training steps) & $15$ \\
         Q-Value clipping & $10.0$ \\
         Munchausen $\tau$ & $0.03$ \\
         Munchausen $\alpha$ & $0.9$ \\
         Munchausen Clipping & $-1$ \\
         Evaluation frequency (w.r.t. dialog turns) & $10000$ \\
         Evaluation dialogs & $500$
    \end{tabular}
    \caption{RL Training Parameters}
\end{table}

\newpage
\section{Dialog Corpus Examples}
\label{sec:dialog_examples}

The following show full examples of dialogs from the corpus. Here we show two dialogs for each goal type and from each type of system.
We show two dialogs for each chosen goal (each from the same type of dialog system) to demonstrate how different users tried to interact with an information seeking dialog agent.

\subsection{Open Goal - CTS Agent}
\subsubsection{Ex. 1}
\begin{dialogue}
    \speak{DIALOG} USER: 6d91b06047716a0a0b2c2cd86fdc78 | POLICY: cts | GOAL-TYPE: OPEN
    \speak{GOAL TEXT} You want to know how to book a hotel
    \speak{USER} book a trip
    \speak{SYSTEM} Are you going on an intracity trip or a business trip?
    \speak{USER} business trip
    \speak{SYSTEM} Did you get written permission from your supervisor?
    \speak{USER} yes
    \speak{SYSTEM} What country are you traveling to?
    \speak{USER} USA
    \speak{SYSTEM} What city are you traveling to?
    \speak{USER} LA
    \speak{SYSTEM} For all  business trips (even for short trips or ones lasting only a single day) outside the country to EU-countries, Iceland, Norway, Liechtenstein or Switzerland, you will need to fill out an A1 Form , which must be brought with you when you leave the country. You will need to fill out a new A1 for each business trip you take. Filling the form out once, does not allow it to apply for subsequent trips you take.
    \speak{SYSTEM} Please check the current COVID-19 travel warnings travel restrictions from the foreign ministry and the RKI. Business trips to high risk areas or virus variation areas are not generally not allowed. In In extreme cases, authorization can be given by the leadership of Department 4 (Administrative Department).
    \speak{SYSTEM} COVID-19: Business trips should be reduced to an absolute minimum and are only allowed when they are absolutely necessary. The decision for necessity can be made by the leaders of the administrative department (see Hygiene concept, 2.4).
    \speak{SYSTEM} Are you planning to extend your trip with private travel?
    \speak{USER} no
    \speak{SYSTEM} Which topic do you want to know more about?
    \speak{USER} accommodation
    \speak{SYSTEM} If the trip is being paid for by a third party (not from the university) or was offered to be paid for by one, no money can be reimbursed.
    \speak{SYSTEM} Where do you plan to stay?
    \speak{USER} hotel
    \speak{SYSTEM} It is required that the bill is addressed to your employer with your name entered as the guest.
    \speak{SYSTEM} In the United States you have a budget of up to 90€ (or 120€ in cities with a population of more than 100,000 or generally/seasonally high hotel prices) for accommodations, so long as you present proof of a paid invoice.
    \speak{SYSTEM} Exceeding the accommodation allowance can be allowed on a case-by-case basis when justified as necessary, e.g., If there were no cheaper hotels near the event location A cheaper hotel farther away would have resulted in higher transportation costs.
    \speak{SYSTEM} Resort taxes may be applied to trips outside of Germany.
    \speak{DIALOG END} SUCCESS
    \speak{DIALOG LENGTH} 16
    \speak{SUBJECTIVE QUALITY} 2
    \speak{SUBJECTIVE LENGTH} 3
\end{dialogue}

\subsubsection{Ex. 2}
\begin{dialogue}
    \speak{DIALOG} USER: d7ea9d8d7835bb048db356e91cfce9 | POLICY: cts | GOAL-TYPE: OPEN
    \speak{GOAL TEXT} You want to know how to book a hotel
    \speak{USER} Tell me how to book a hotel
    \speak{SYSTEM} What country are you traveling to?
    \speak{USER} USA
    \speak{SYSTEM} Are you planning to extend your trip with private travel?
    \speak{USER} No
    \speak{SYSTEM} Hotels can be booked using the university's partner travel agency, through a booking portal, or booked directly The contact for the partner travel agency can be found here. The university recommends these hotels.
    \speak{DIALOG END} SUCCESS
    \speak{DIALOG LENGTH} 4
    \speak{SUBJECTIVE QUALITY} 3
    \speak{SUBJECTIVE LENGTH} 3
\end{dialogue}

\subsection{Easy Goal - FAQ Agent}
\subsubsection{Ex. 1}
\begin{dialogue}
    \speak{DIALOG} USER: 7addf36ad46517524764cd73b84d5b | POLICY: faq | GOAL-TYPE: EASY
    \speak{GOAL TEXT} You want to know if you can be reimbursed if you need to book a taxi during your trip.
    \speak{USER} If I need to book a taxi during my trip, can I be reimbursed? If so, how should I do this?
    \speak{SYSTEM} A taxi is only reimbursable if there is a business related reason or a pressing personal reason, e.g.: insufficient public transportation, Health reasons, Disability.
    \speak{DIALOG END} SUCCESS
    \speak{DIALOG LENGTH} 1
    \speak{SUBJECTIVE QUALITY} 3
    \speak{SUBJECTIVE LENGTH} 2
\end{dialogue}

\subsubsection{Ex. 2}
\begin{dialogue}
    \speak{DIALOG} USER: 266f4afc0f4957543ecd710668a801 | POLICY: faq | GOAL-TYPE: EASY
    \speak{GOAL TEXT} You want to know if you can be reimbursed if you need to book a taxi during your trip.
    \speak{USER} hello
    \speak{SYSTEM}Thank you for using the business travel chatbot.Please let us know about your experience by filling out a quick survey, that way we can improve our performance and your experience.
    \speak{} RESET
    \speak{USER} hello i would like some information please
    \speak{SYSTEM} Please use the country database  to inform yourself about possible  Risks related to your trip. It is required to confirm that you have checked this information and gotten necessary advice as part of your travel application.  Your supervisor will confirm this when signing the form giving you permission to travel.
    \speak{DIALOG END} FAILURE
    \speak{DIALOG LENGTH} 1
    \speak{SUBJECTIVE QUALITY} 1
    \speak{SUBJECTIVE LENGTH} 1
\end{dialogue}

\subsection{Hard Goal - HDC Agent}
\subsubsection{Ex. 1}

\begin{dialogue}
    \speak{DIALOG} USER: ba401ddeae9aef6345d8c3dbddb3bf | POLICY: hdc | GOAL-TYPE: HARD
    \speak{GOAL TEXT} You want to know how much money you can get for your accommodations. <ul><li>You are traveling to France for your next trip</li><li>You plan to stay with your brother in his apartment. </li></ul>
    \speak{USER} Book a trip
    \speak{SYSTEM} Are you going on an intracity trip or a business trip?
    \speak{USER} what is the difference between an intracity business trip and business trip?
    \speak{SYSTEM} Business trips are business related travel outside of the city. The following are not business trips: Company outings, Trips for a side job, Travel to a continuing education training (if it is done out of private interest)
    \speak{SYSTEM} Intracity business trips are business related travel within the city.
    \speak{} RESET
    \speak{USER} Book a trip
    \speak{SYSTEM} Are you going on an intracity trip or a business trip?
    \speak{USER} intracity business trip
    \speak{SYSTEM} Did you get verbal permission from your supervisor?
    \speak{USER} yes
    \speak{SYSTEM} Do you want to know more about booking Transportation or about ancillary costs?
    \speak{USER} incidental costs
    \speak{SYSTEM} Which incidental costs ocurred during your trip or do you anticipate occurring
    \speak{USER} accomodation
    \speak{SYSTEM} Meals which are part of a conference are reimbursable, as long as the conference is being attended out of business interests.
    \speak{} RESET
    \speak{USER} book a trip
    \speak{SYSTEM} Are you going on an intracity trip or a business trip?
    \speak{USER} business trip
    \speak{SYSTEM} Did you get written permission from your supervisor?
    \speak{USER} yes
    \speak{SYSTEM} What country are you traveling to?
    \speak{USER} (PRE-NLU) France
    \speak{SYSTEM} What city are you traveling to?
    \speak{USER} Paris
    \speak{SYSTEM} For all  business trips (even for short trips or ones lasting only a single day) outside the country to EU-countries, Iceland, Norway, Liechtenstein or Switzerland, you will need to fill out an A1 Form , which must be brought with you when you leave the country. You will need to fill out a new A1 for each business trip you take. Filling the form out once, does not allow it to apply for subsequent trips you take.
    \speak{SYSTEM} Please check the current COVID-19 travel warnings travel restrictions from the foreign ministry and the RKI. Business trips to high risk areas or virus variation areas are not generally not allowed. In In extreme cases, authorization can be given by the leadership of Department 4 (Administrative Department).
    \speak{SYSTEM} COVID-19: Business trips should be reduced to an absolute minimum and are only allowed when they are absolutely necessary. The decision for necessity can be made by the leaders of the administrative department (see Hygiene concept, 2.4).
    \speak{SYSTEM} Are you planning to extend your trip with private travel?
    \speak{USER} yes
    \speak{SYSTEM} (A private extension to a trip must remain in proportion to the length of business-related travel. Business interests must remain the priority. For questions, please contact the travel department.
    \speak{SYSTEM} Which topic do you want to know more about?
    \speak{USER} accomodation
    \speak{SYSTEM} If the trip is being paid for by a third party (not from the university) or was offered to be paid for by one, no money can be reimbursed.
    \speak{SYSTEM} Where do you plan to stay?
    \speak{USER} with family
    \speak{SYSTEM} Outside of Germany, you are entitled to 30€ if you are staying with friends or family.
    \speak{DIALOG END} SUCCESS
    \speak{DIALOG LENGTH} 14
    \speak{SUBJECTIVE QUALITY} 4
    \speak{SUBJECTIVE LENGTH} 4
\end{dialogue}

\subsubsection{Ex. 2}
\begin{dialogue}
    \speak{DIALOG} USER: 3531297f0bd32d54dcc23e8e8a50ad | POLICY: hdc | GOAL-TYPE: HARD
    \speak{GOAL TEXT} You want to know how much money you can get for your accommodations. <ul><li>You are traveling to France for your next trip</li><li>You plan to stay with your brother in his apartment. </li></ul>
    \speak{USER} how much am i entitled to claim for accomodation
    \speak{SYSTEM} What emergency are you experiencing?
    \speak{USER} not an emergency, but accommodation costs
    \speak{SYSTEM} What do I do in an emergency on a business trip? Please call the number listed on the Emergency-Card :  +123456789. The Emergency-Card is also available in business card format. For more information, please enquire with the business travel department or your institute. This phone number is for emergencies only! Do not distribute it. Emergency App24/7 Emergency calling, Text chat, Video advice per the Emergency call Business App for iOS and AndroidDownload link: Notfall-AppQR-Code:Registration code: Note: distributing the registration code to third parties is strictly prohibited.
    \speak{SYSTEM} Do you have any further questions?
    \speak{USER} yes
    \speak{SYSTEM} What topic do you have questions about? You can either click on an answer from the suggested topics or enter your own text.
    \speak{USER} how much can i get back for accommodation costs?
    \speak{DIALOG END} FAILURE
    \speak{DIALOG LENGTH} 5
    \speak{SUBJECTIVE QUALITY} 1
    \speak{SUBJECTIVE LENGTH} 1    
\end{dialogue}

\newpage
\section{Data Agreement}
Before beginning the experiment, users were provided with a data agreement. 
Although we did not collect any personally identifying data, we wanted to make sure that users were aware of what they would be asked to do, the purpose of the research, what data we would collect and how the data would be processed. 

\begin{figure}[!htb]
    \centering
    \includegraphics[width=\textwidth]{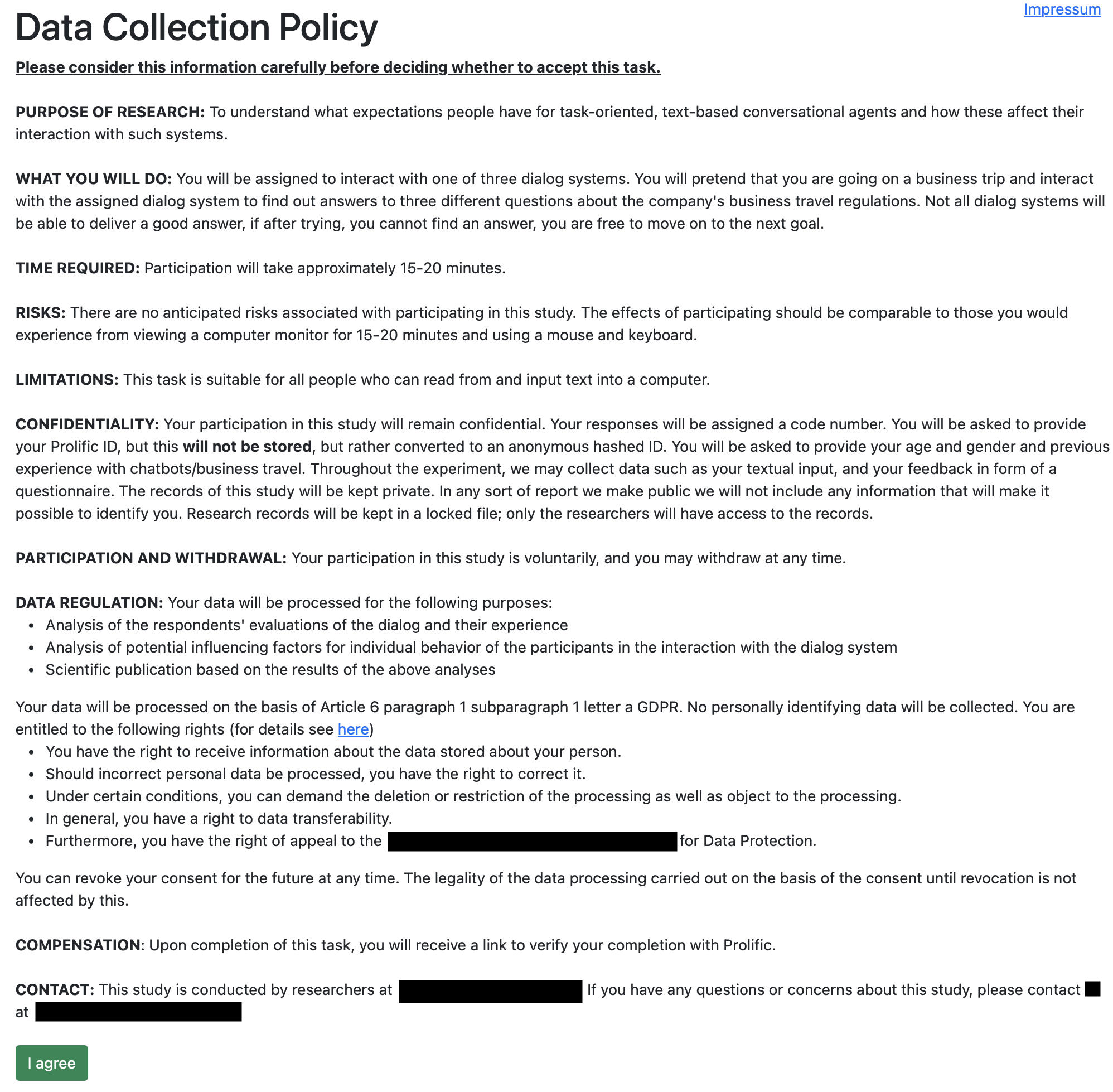}
\end{figure}

\newpage
\section{Study Instructions}
During the interaction, users were provided with the following interface, on the right side they had an information goal for which they should find an answer.
On the left side, they had a window with their conversation with the chatbot.
Once they felt they had found an answer to their question, they could click on the button underneath the goal to move on to the next dialog.

\begin{figure}[!htb]
    \centering
    \includegraphics[width=\textwidth]{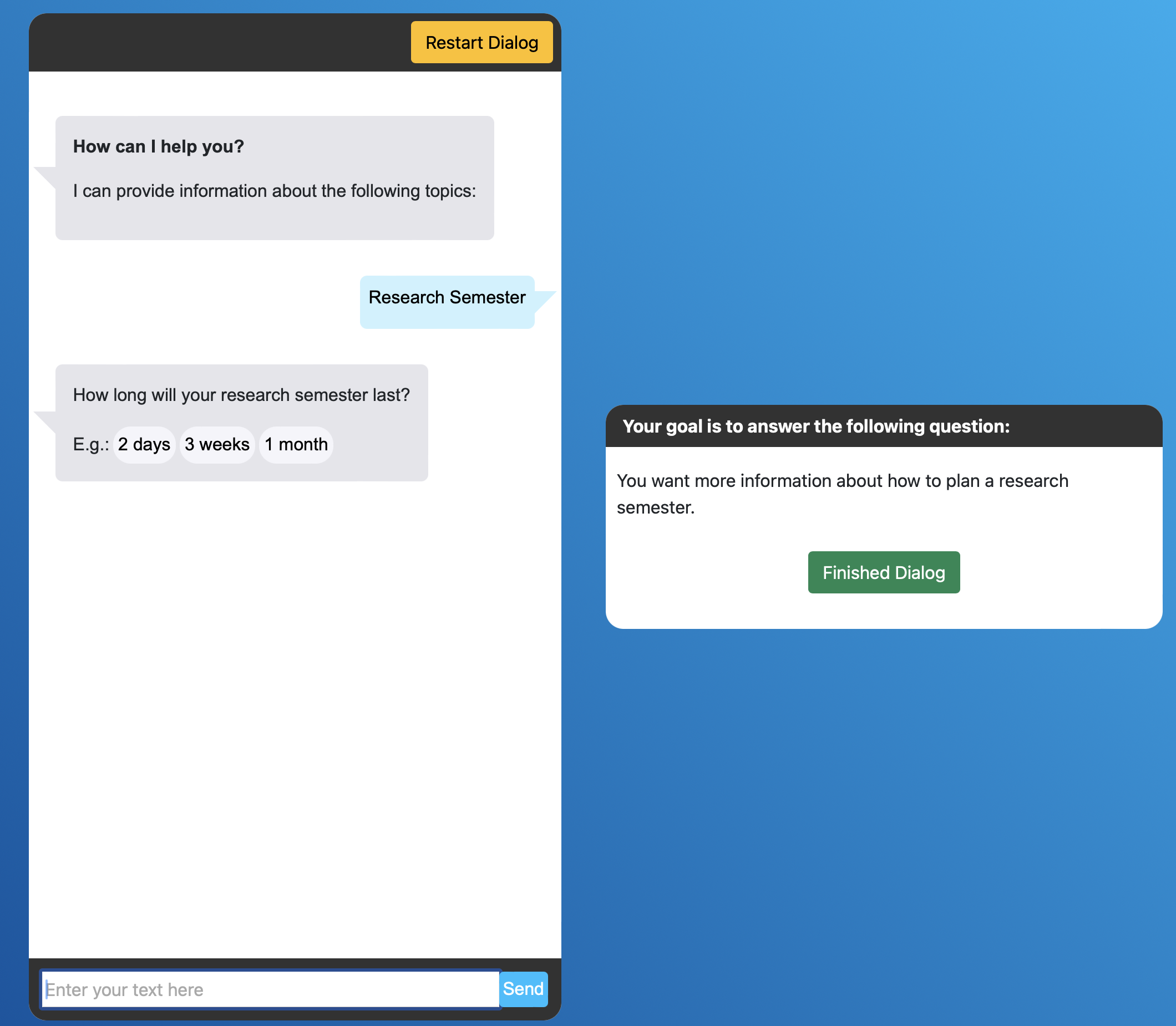}
\end{figure}

\newpage
\section{Interaction Surveys}
\label{sec:surveys}

\subsection{Pre-Interaction Survey}
The survey given to users before the interaction can be seen below. 
Here they were asked general questions about their demographics, previous experience with the domain and chatbots.
Users were also asked Likert and free-response questions about their expectations for an information seeking chatbot. In particular about how they expected to be able to input text and and how they expected the chatbot to answer.

\begin{figure}[!htb]
    \centering
    \includegraphics[width=\textwidth]{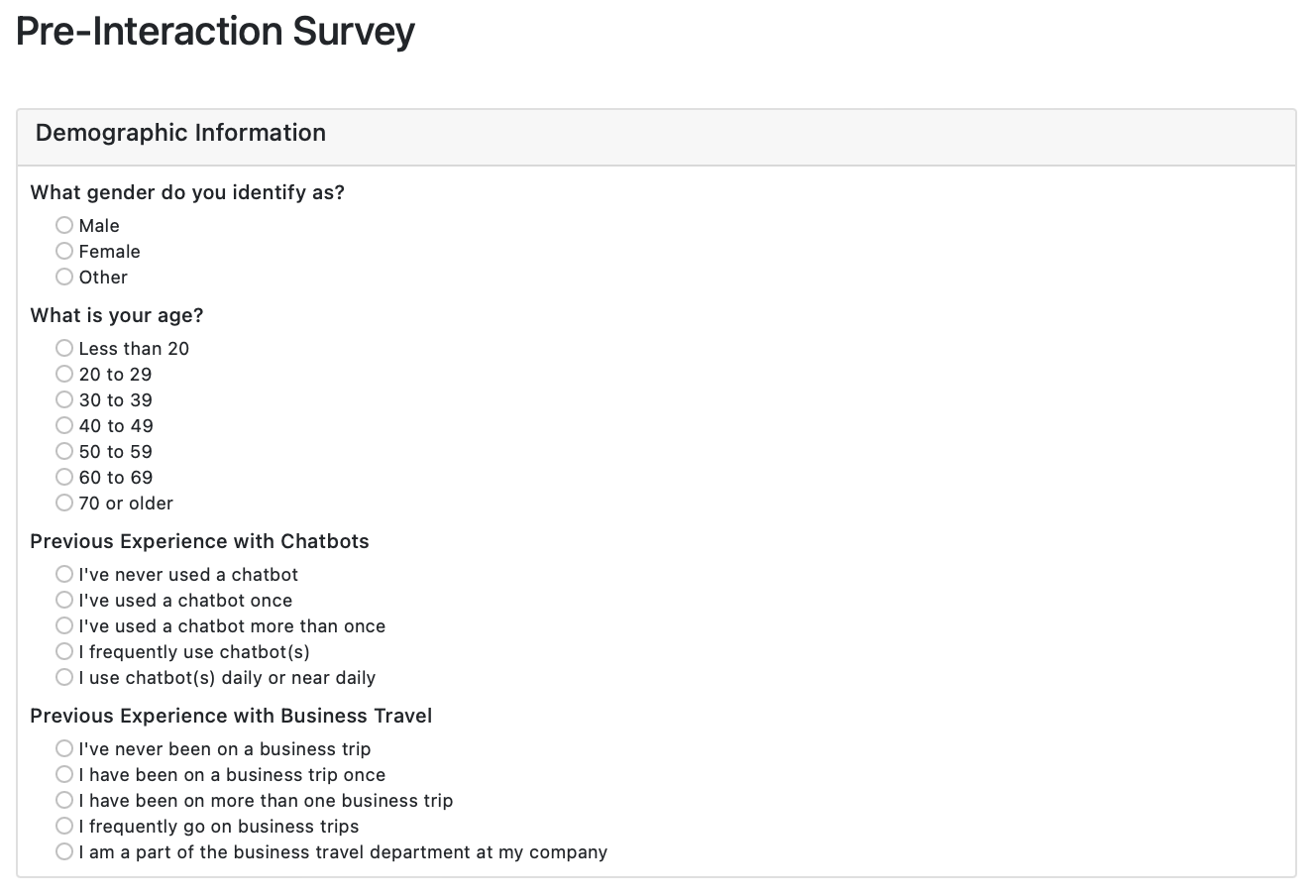}
\end{figure}

\begin{figure}[!htb]
    \centering
    \includegraphics[width=\textwidth]{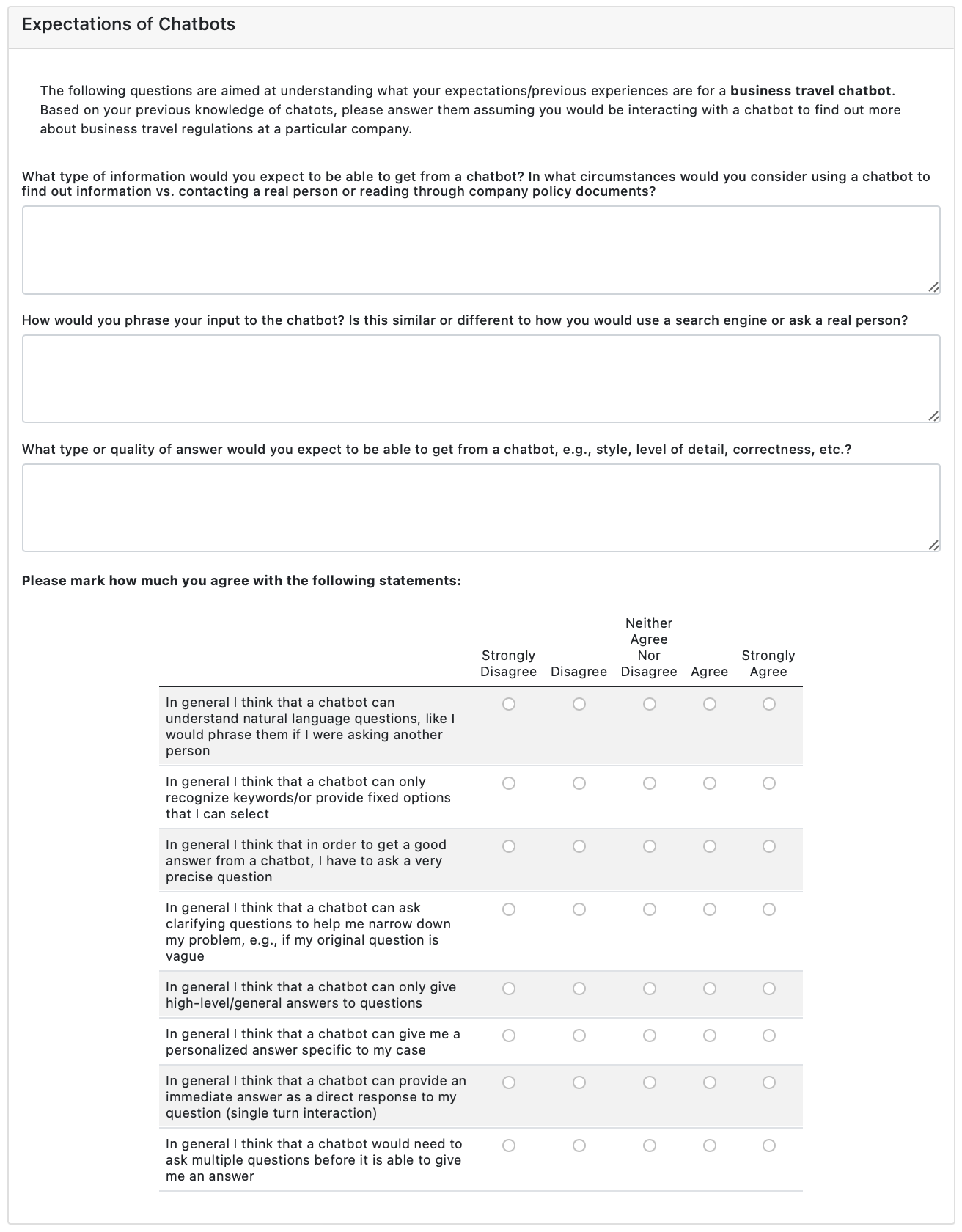}
\end{figure}

\newpage
\subsection{Post-Dialog Survey}

After each interaction, users were asked to rate their perception of the dialog length on a five-point Likert scale and their perception of how well their question was answered on a four-point Likert scale.

\begin{figure}[!htb]
    \centering
    \includegraphics[width=0.75\textwidth]{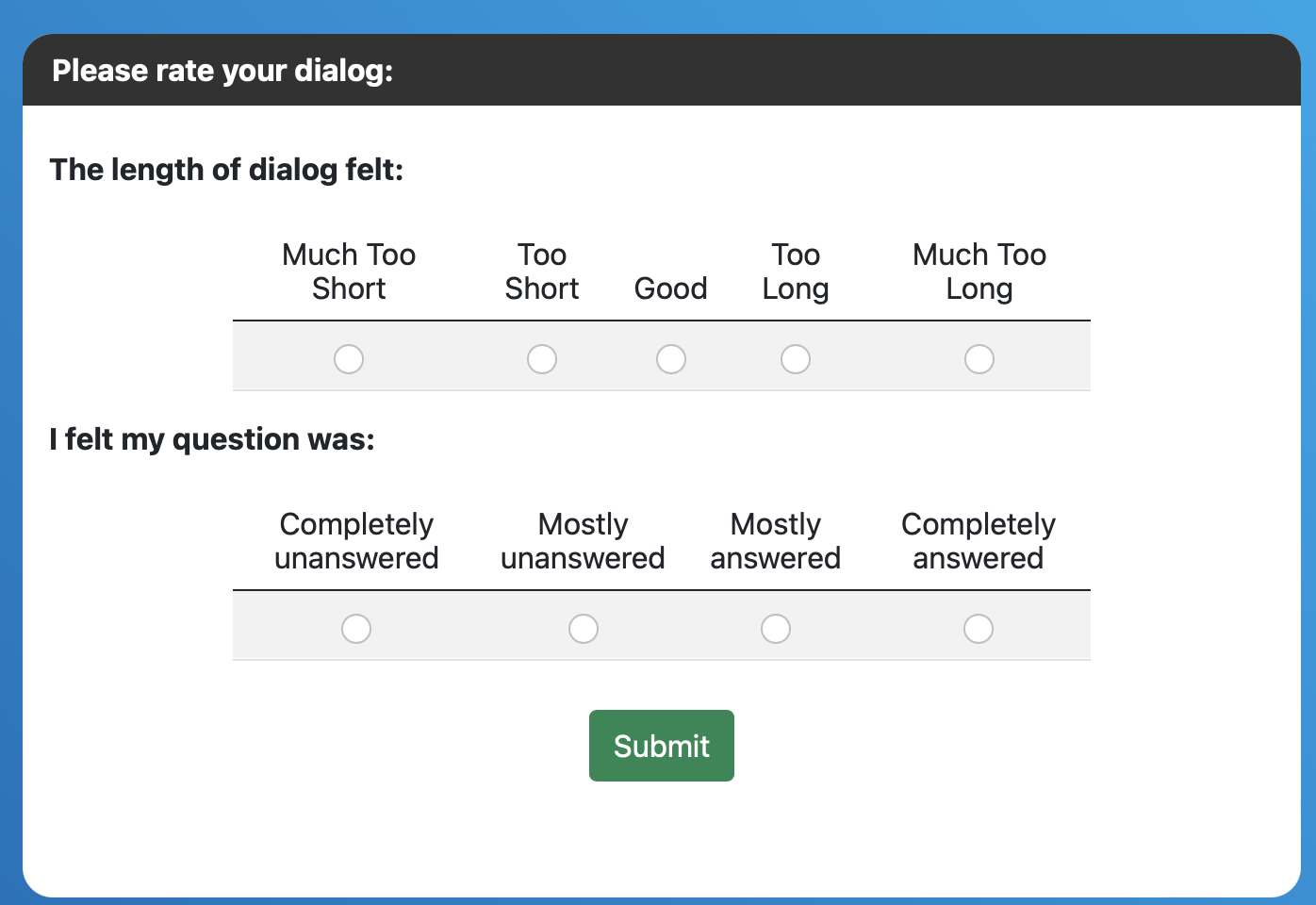}
\end{figure}

\newpage
\subsection{Post-Interaction survey}

The survey given to users after interacting with their assigned style of chatbot can be seen below.
Users were asked to provide free-form feedback about how well their interactions were met as well as to answer the same Likert questions about their mental model of a chatbot that they had answered prior to the interaction.
Users were also asked to fill out a usability questionnaire \citep{finstad2010usability}
and the trust and reliability subscales from the Trust in Automation questionnaire \citep{korber2018theoretical} as well as answering free-form questions on their experience and perception of the chatbot.

\begin{figure}[!htb]
    \centering
    \includegraphics[width=\textwidth]{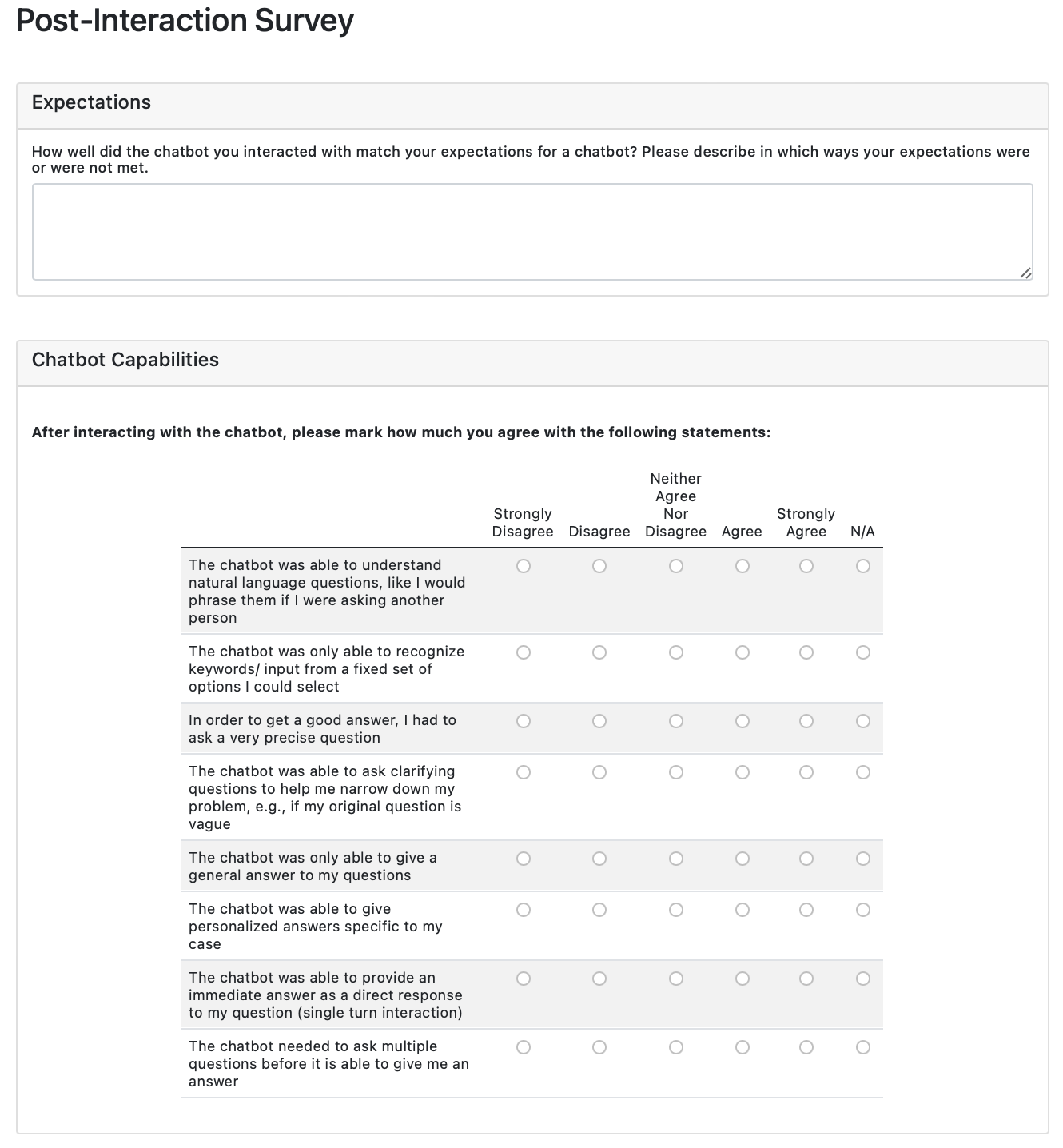}
\end{figure}

\begin{figure}[!htb]
    \centering
    \includegraphics[width=\textwidth]{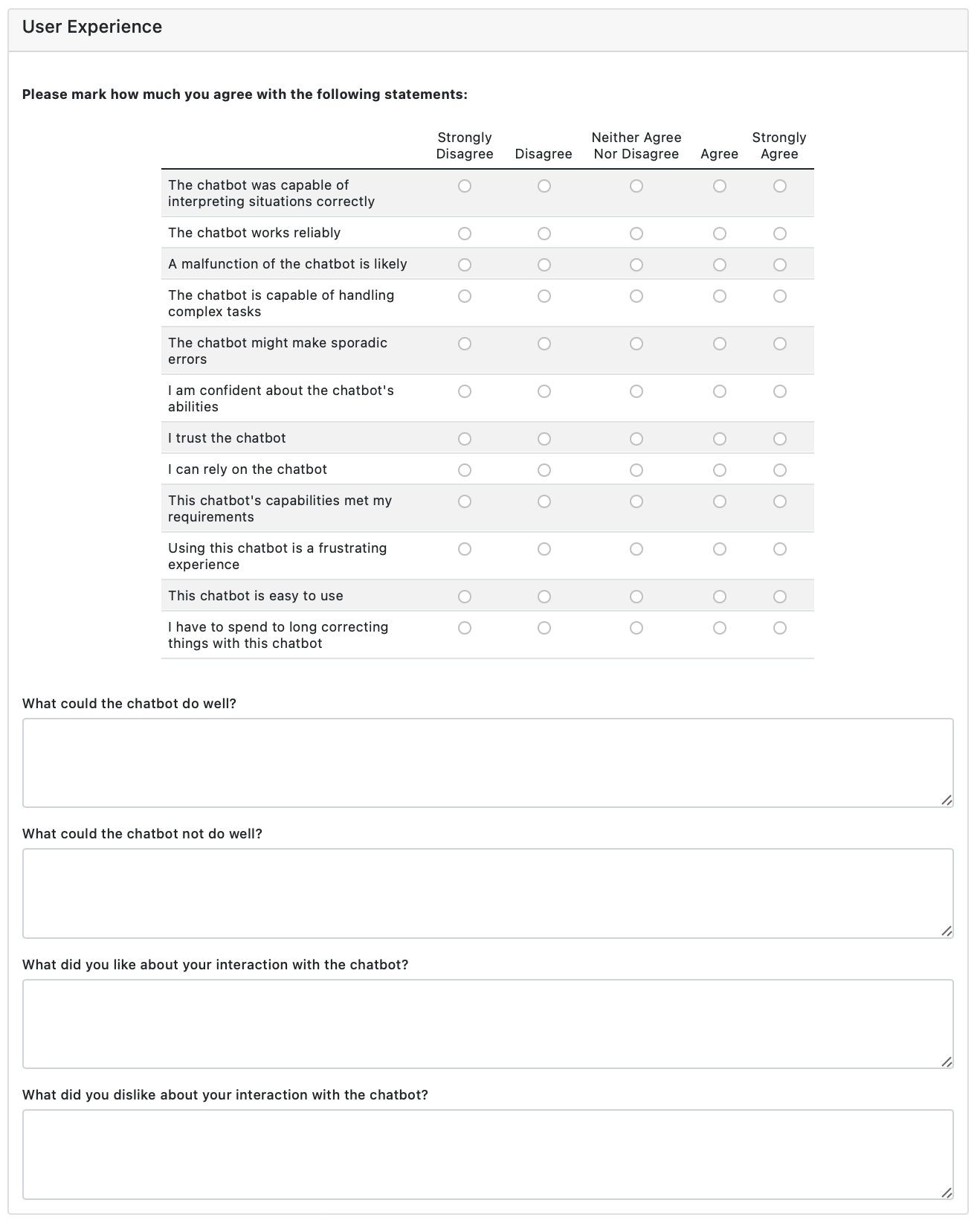}
\end{figure}

\end{document}